\documentclass{article}

\usepackage{main}

\usepackage[colorlinks=true,citecolor=blue]{hyperref}
\usepackage[english]{babel}
\usepackage{graphicx}
\usepackage{amsmath}
\usepackage{amsfonts}
\usepackage{amssymb}
\usepackage{mathtools}
\usepackage[numbers]{natbib}
\usepackage{bm}
\usepackage{mathtools}
\usepackage{algorithm}
\usepackage[noend]{algpseudocode}
\usepackage{amsthm}

\newtheorem{theorem}{Theorem}[section]
\newtheorem{example}{Example}[section]
\newtheorem{definition}{Definition}[section]
\newtheorem{proposition}{Proposition}[section]

\theoremstyle{definition}
\DeclareMathOperator*{\argmin}{\arg\!\min}
\DeclareMathOperator*{\argmax}{\arg\!\max}

\title{Stacking and stability}

\author{
  Nino Arsov \\
  Macedonian Academy of Sciences and Arts\\
  1000 Skopje, Macedonia \\
  \texttt{narsov@manu.edu.mk} \\
   \And
 Martin Pavlovski \\
  Temple University\\
  Philadelphia, PA 19122 \\
  \texttt{martin.pavlovski@temple.edu} \\
  \And
  Ljupco Kocarev \\
  Macedonian Academy of Sciences and Arts\\
  1000 Skopje, Macedonia\\
  \texttt{lkocarev@manu.edu.mk}
}

\begin{document}

\maketitle

\begin{abstract}
Stacking is a general approach for combining multiple models toward greater predictive accuracy. It has found various application across different domains, ensuing from its meta-learning nature. 
Our understanding, nevertheless, on how and why stacking works remains intuitive and lacking in theoretical insight. In this paper, we use the stability of learning algorithms as an elemental analysis framework suitable for addressing the issue.
To this end, we analyze the hypothesis stability of stacking, bag-stacking, and dag-stacking and establish a connection between bag-stacking and weighted bagging. We show that the hypothesis stability of stacking is a product of the hypothesis stability of each of the base models and the combiner. Moreover, in bag-stacking and dag-stacking, the hypothesis stability depends on the sampling strategy used to generate the training set replicates.
Our findings suggest that 1) subsampling and bootstrap sampling improve the stability of stacking, and 2) stacking improves the stability of both subbagging and bagging.

\end{abstract}

\section{Introduction}
\label{intro}

Stacked generalization (stacking)~\citep{wolpert1992stacked} is a prominent and popular off-the-shelf meta-learning approach suitable for a variety of downstream machine learning tasks in areas such as computer vision and computational biology. Stacking is related to ensemble learning approaches such as bagging~\citep{breiman1996bagging} and boosting~\citep{freund1995desicion}.
With the rising prevalence of machine learning techniques for everyday problem solving, it has also reached the status of a missing-ingredient algorithm in competition platforms such as Kaggle. Moreover, stacking has recently found various applications across different domains~\citep{anifowose2015improving,yu2015learning,nicolas2015adaptive,xing2016temporal,bhatt2017improved,healey2018mapping,yao2018using, malmasi2018native, xia2018novel, peyret2018automatic, wang2018significantly}.

Currently, however, there is still very little insight into how and why stacking works, barring rare blog posts by machine learning practitioners in which the explanations tend to be facile, based on intuition, and lacking in theoretical insight.
In the machine learning research community, there are only two scant papers that investigate the effectiveness of stacking and its variations,  called bag-stacking and dag-stacking, in terms of predictive accuracy~\citep{ting1997stacked,bag_stacking}.

A common way to analyze a learning algorithm's capability to generalize is to see how sensitive it is with respect to small changes in the training set. Sensitivity can be quantified using different notions of stability, and once we know how stable a learning algorithm is, we can use that to derive an upper bound on its generalization error. 
The research focused on bounding the generalization error has been prolific and has produced significant advances toward understanding how good can learning algorithms generalize.  But, more importantly, it has established a precise relationship between stability and generalization.
There are different classes of upper bounds that vary depending on how tight they are; tighter upper bounds convey more reliable estimates of the generalization error. Examples include upper bounds based on the Vapnik-Chervonenkis (VC) dimension~\citep{vapnik1999overview}, probably approximately correct (PAC) learning~\citep{valiant1984theory}, PAC-Bayes bounds~\citep{ambroladze2007tighter}, the stability of learning algorithms~\citep{elisseeff1, elisseeff_randomized}, and so on.

Inspired by upper bounds as a tool to assess the generalization error of a learning algorithm, we formally investigate how stability interacts in stacking, bag-stacking, and dag-stacking. This paper is a first look into the effectiveness of these three approaches from the perspective of stability analysis.
The contributions of this paper are: 

\begin{itemize}
    \item[$\bullet$] we analyze the hypothesis stability of stacking, bag-stacking, and dag-stacking,
    \item[$\bullet$] we establish a connection between bag-stacking and weighted bagging, i.e., we show that bag-stacking equals weighted bagging,
    \item[$\bullet$] we show that subsampling/bootstrap sampling improves the hypothesis stability of stacking and that stacking improves the hypothesis stability of both dagging (subbagging) and bagging.
\end{itemize}
Regarding the third contribution, we additionally show that stacking improves the stability of the stacked learning algorithms by a factor of $1/m$, whereas using bootstrap sampling lends further improvements on top of stacking.

In this paper, we float a thorough description of the way stability interacts in bagging, stacking, and in-between, i.e., in bag-stacking. We maintain that stability is the appropriate formal apparatus to demonstrate the effectiveness of stacking toward reducing the generalization error of the base models.
We show that, in contrast to the traditional heterogeneous ensemble structure of stacking, its homogeneous counterpart, bag-stacking, can be used instead to stabilize the algorithm further and, consequently, yield a tighter stability-based upper bound, reducing the generalization error further.

The paper is organized in the following sections: Section~\ref{sec:notation} introduces some definitions and the notation used throughout the text, Section~\ref{sec:1} discusses the link between meta-learning and ensemble learning approaches, Section~\ref{sec:2} presents the existing work on the stability of deterministic and randomized learning algorithms, and Section~\ref{sec:main} contains the main results and contributions. We conclude the paper with Section~\ref{sec:conclusion}.

\section{Notation and Prerequisites}
\label{sec:notation}

This section briefly introduces the notation used in the rest of the paper and the prerequisite notions that  help analyze learning algorithms from the perspective of stability.
Bold symbols represent vectors, such as $\bm{x}\in\mathbb{R}^3$, whereas $x\in\mathbb{R}$ is a scalar. The symbols below designate various quantities or measures.
\begin{itemize}
\item[$\bullet$] $\mathbb{P}[\cdot]$: probability
\item[$\bullet$] $\mathbf{E}_X[\cdot]$: expected value with respect to $X$
\item[$\bullet$] $\text{Var}(\cdot)$: variance
\item[$\bullet$] $e(\cdot)$: error function
\end{itemize}

Let $\mathcal{X}$ and $\mathcal{Y}$ denote the input and output space of a learning algorithm. Here, we assume that $\mathcal{X} \subseteq V^d$, where $d \geq 1$ and $V$ is a vector space, while $\mathcal{Y} \subseteq \mathbb{R}$ for regression tasks, and $\mathcal{Y}=\mathcal{C}$, which is a set of labels, not necessarily numerical, for classification tasks.
\\
Let $\mathcal{Z}=\mathcal{X} \times \mathcal{Y}$. Consider a sample $\mathcal{D}$ of $m$ input-output pairs,
\begin{equation*}
\mathcal{D}=\{z_1=(\bm{x}_1,y_1),z_2=(\bm{x}_2,y_2),\ldots , z_m=(\bm{x}_m,y_m) \}=(\mathcal{D}_X, \mathcal{D}_Y).
\end{equation*}
The examples $z_1, z_2, \ldots z_m$ are drawn i.i.d. from an unknown distribution $P$ and comprise the \textit{training set} $\mathcal{D}$. If we take $\mathcal{Z}^m$ to denote the set of all possible training sets of size $m$, then $\mathcal{D}\in\mathcal{Z}^m$. 
By removing the $i$-th example from $\mathcal{D}$,  we get a new training set of $m-1$ examples,
\begin{equation*}
\mathcal{D}^{\backslash i} = \{z_1,\ldots,z_{i-1},z_{i+1},\ldots,z_m\}.
\end{equation*}
Replacing the $i$-th example from $\mathcal{D}$ with some $z \in\mathcal{Z}$ and $z \notin \mathcal{D}$ yields
\begin{equation*}
\mathcal{D}^{\backslash i} \cup \{z\} = \{z_1,\ldots,z_{i-1},z,z_{i+1},\ldots,z_m\},\quad z \in \mathcal{Z},z \notin \mathcal{D}.
\end{equation*}

A training algorithm $A$ learns a mapping (function) $f:\mathcal{X}\rightarrow\mathcal{Y}$ that approximates $P$. When the learning algorithm $A$ learns $f$ using a training set $\mathcal{D}$, we denote this by $f_{A(\mathcal{D})}$. For brevity, however, it is useful to omit $A$ and reduce $f_{A(\mathcal{D})}$ to $f_{\mathcal{D}}$, barring a few cases when it is necessary to distinguish different learning algorithms. In addition, $f_{\mathcal{D}}$ also refers to a \textit{model} (for $f$), a \textit{hypothesis}, or a \textit{predictor}.
Lastly, for any $\mathcal{S} \subseteq \mathcal{Z}$, we define
\begin{equation*}
f_{\mathcal{D}}(\mathcal{S}_X) = \{ f_{\mathcal{D}}(\bm{x})\,|\,\bm{x}\in\mathcal{S}_X\},
\end{equation*}
where $\mathcal{S}_Y$ and $f_{\mathcal{D}}(\mathcal{S}_X)$ retain the order of examples.

\section{Combining Predictors: Ensemble Learning}
\label{sec:1}

Ensemble learning is a modeling paradigm in which at least two predictors are combined into a single model, called an \textit{ensemble}. At its simplest, an ensemble is a merger of \textit{weak} models having a wanting capability to generalize. Combined, they generalize significantly better.

In 1988, Kearns and Valiant introduced a riddle that asked whether weak models could turn into a single strong predictor.
Weak models are considered to predict slightly better than random guessing, and a succinct definition can be found in~\citep[Def. 4.1]{kutin2001interaction}. A paper by Rob Schapire analyzed the predictive strength of weak models as early as 1990~\citep{schapire1990strength}.
Ensemble learning has seen two major breakthroughs: bootstrap aggregation (bagging)~\citep{breiman1996bagging} and boosting~\citep{freund1995desicion}. Both are meta-learning approaches that aim to ``boost'' the accuracy of a given algorithm. Moreover, bagging is designed to reduce \textit{variance}, whereas boosting reduces both \textit{bias} and \textit{variance} that comprise the expected prediction error $e=\text{bias} + \text{variance}^2$.

\subsection{Bagging as a parallel and independent ensemble learning approach}
In bagging, a learning algorithm repeatedly trains weak models using different bootstrap samples $\mathcal{D}_1,\mathcal{D}_2,\ldots,\mathcal{D}_T$ drawn from a training set $\mathcal{D}$.  The training rounds are independent and thus easily parallelizable. The trained models are then fused into a single predictor $f_a$ by aggregation. Bagging does not minimize a particular loss function for it is engineered to reduce variance and stabilize an unstable learning algorithm by aggregation~\citep{breiman1996bagging}.
The basic reasoning behind bagging is that we want to train weak learners using many different training sets from $\mathcal{Z}^m$ and then aggregate them by taking their expected value
\begin{equation}
    f_a=\mathbf{E}_{\mathcal{D}}[f_{\mathcal{D}}].
    \label{eq:agg_bagging}
\end{equation} In reality, however, the data distribution $P$ is unknown and there is only one training set $\mathcal{D}$ available to work with. Bootstrap sampling, or random uniform sampling with replacement, thus allows one to repeatedly re-sample $\mathcal{D}$ and generate many different replicates of the training set in order to simulate and estimate the aggregated predictor $f_a$ by $F_{\mathcal{D}}$: 

\begin{equation*}
\hat{f}_a(\bm{x})=F_{\mathcal{D}}(\bm{x})=\frac{1}{T}\sum_{t=1}^T f_{\mathcal{D}_t}(\bm{x})
\end{equation*}

A large number of replicates gives a better estimate of $f_a=\mathbf{E}_{\mathcal{D}}[f_{\mathcal{D}}]$ since the sample used to estimate $f_a$ is larger.
Although the error of $f_a$ does not taper off as the number of replicates increases, its improvement starts faltering at some point. The reason being is that when the number of bootstrap samples becomes sufficiently large to provide a very accurate estimate of $f_a$, adding new samples makes only minute improvements. The bagging algorithm is given in Algorithm~\ref{alg:bagging}.


\begin{algorithm}
\caption{Bagging~\citep{breiman1996bagging}}
\label{alg:bagging}
\begin{algorithmic}[1]
\Procedure{Bagging}{$A, \mathcal{D}, T$}
\For{$t=1,\dots,T$}
\State $\mathcal{D}_t \gets BootstrapSampling(\mathcal{D})$\Comment{Generate a bootstrap sample}
\State $f_{\mathcal{D}_t}^{(t)} \gets A (\mathcal{D}_t)$\Comment{Apply $A$ on $\mathcal{D}_t$}
\EndFor
\State \textbf{return} $F_{\mathcal{D}}(\bm{x})=\frac{1}{T}\sum_{t=1}^T f_{\mathcal{D}_t}(\bm{x})$\Comment{For regression}
\State \phantom{\textbf{return}} $F_{\mathcal{D}}(\bm{x})=\argmax_{\mathcal{C}_k} \sum_{t=1}^T [I(f_{\mathcal{D}_t}(\bm{x}) = \mathcal{C}_k)]_{k=1}^K$\Comment{For classification}
\EndProcedure
\end{algorithmic}
\end{algorithm}

\noindent
A variation of bagging, called subbagging, is based on sampling from $\mathcal{D}$ \textit{without} replacement to generate the $T$ training sets, each comprising $n < m$ examples. Although this might compound stability in contrast to bagging, it speeds up the computation when $n$ is significantly smaller than $m$. The stability analysis is also deferred to Section~\ref{sec:main}.

\subsection{Boosting as a sequential and dependent ensemble learning approach}

Boosting~\citep{freund1995desicion} follows the same ensemble learning paradigm as bagging, but it has two key differences.
By contrast, it minimizes the exponential loss $e^{yF(\bm{x})}$, which is an upper bound on the $\{0,1\}$ classification loss. In the first round, all training examples are equally important.
In each following round, the algorithm adjusts the importance so that the current weak model focuses on examples mispredicted by the preceding one, while those already learned become less important. This way, the boosting algorithm transforms the training set $\mathcal{D}$ into $T$ different training sets $\mathcal{D}(\bm{w}_1), \mathcal{D}(\bm{w}_2),\ldots, \mathcal{D}(\bm{w}_T)$, where $\bm{w}_t\in\mathbb{R}^m$ and $\|\bm{w}_t\|_1=1$ for $t=1,2,\ldots,T$. These imporatnces form a probability distribution, i.e., they add up to 1. They have the same effect as oversampling $\mathcal{D}$.  The least important example occurs once in the oversampled set, while the rest occur multiple times. This approach, unlike bagging, introduces a dependency between the weak models.
At the same time, the $t$-th weak learner gets a weight $\alpha_t$ reflecting its error. Accurate weak models get higher weights that are then used to aggregate them into a single strong model with boosted accuracy using the mixture

\begin{equation*}
\hat{f}_a(\bm{x})=F_{\mathcal{D}}(\bm{x}) = \sum_{t=1}^T \alpha_t f_{\mathcal{D}_t}(\bm{x}).
\end{equation*}

The second difference is that boosting is a fully deterministic algorithm that minimizes the loss function $e^{yF(\bm{x})}$. In other words, boosting is a method for margin maximization~\citep{schapire1998boosting}. More details on how each step in Algorithm~\ref{alg:boosting} helps minimize the exponential loss can be found in~\cite{freund1995desicion}.

\begin{algorithm}
\caption{AdaBoost.M1~\citep{freund1995desicion}}
\label{alg:boosting}
\begin{algorithmic}[1]
\Procedure{AdaBoost.M1}{$A,\mathcal{D},T$}

\State Initialize the example importances $\omega_{1i} = 1/m$, $i = 1, 2, \dots, m$
\For{$t=1,\dots,T$}
    \State $f_{\mathcal{D}_t} \gets A(\mathcal{D}(\bm{\omega}_t))$\Comment{Apply $A$ on $\mathcal{D}(\bm{\omega}_t)$}
    \State
    \[
    \text{err}_t \gets \frac{\sum_{i=1}^m \omega_{ti} I(y_i \neq f_{\mathcal{D}_t}(x_i))}{\sum_{i=1}^m \omega_{ti}}.
    \]\Comment{Compute the weighted error}
    \State $\alpha_t \gets \log((1 - \text{err}_t) / \text{err}_t)$ \Comment{Compute $f_{\mathcal{D}_t}$'s weight}
    \State $\omega_{ti} \gets \omega_{ti} \exp [\alpha_t I(y_i \neq f_{\mathcal{D}_t}(\bm{x}_i))]$, $i = 1, \dots, m$ \Comment{Adjust the example importances}
\EndFor
\State \textbf{return} $F_{\mathcal{D}}(\bm{x})=\frac{1}{T}\sum_{t=1}^T \alpha_t f_{\mathcal{D}_t}(\bm{x})$\Comment{For regression}
\State \phantom{\textbf{return}} $F_{\mathcal{D}}(\bm{x})=\argmax_{\mathcal{C}_k} \sum_{t=1}^T \alpha_t [I(f_{\mathcal{D}_t}(\bm{x}) = \mathcal{C}_k)]_{k=1}^K$\Comment{For classification}
\EndProcedure
\end{algorithmic}
\end{algorithm}

\subsection{Stacking: a meta-learning perspective}
\label{sec:1.2}
Meta-learning strategies are used to learn latent variables of a model in supervised and unsupervised settings. They always aim to consolidate various virtues of the model. On one hand, meta-learning works with trained models and uses ground truth data, like boosting, to find the optimal mixture weights $\boldsymbol{\alpha}$ to combine weak models.  Regularization, on the other hand, is added to the training process to mitigate overfitting. 
In general, meta-learning approaches operate on top of either the input space, the output space, or the span of the parameter space. They have a vague nature that makes them more involving than classical learning in the sense that they can often be heuristic and allow different interpretations of the latent variables across learning scenarios.

Stacking is a meta-learning approach for combining multiple models. Stacking operates over the output space and is strictly limited to supervised learning settings. At its simplest, it is a two-level pipeline of models, where predictions from the first level move on to the second to act as the input to a separate learning component called a \textit{combiner}. Least-squares and logistic regression are regular instances of combiner algorithms for regression and classification tasks. Stacking has an intrinsic connection to ensemble learning.

Stacking combines $T \geq 2$ models, each of which is trained using a (usually different) learning algorithm over the same training set $\mathcal{D}$. Thus there are $T$ learned mappings from $\mathcal{X}$ to $\mathcal{Y}$ that we denote by $f_{\mathcal{D}}^{(t)}$, $t=1,2,\ldots,T$. Adequately, there are $T$ sets of predictions on $\mathcal{D}$, $f_{\mathcal{D}}^{(t)}(\mathcal{D}_X),\, t=1,2,\ldots,T$.
The combiner learns a new mapping $g$ with unknown parameters $\boldsymbol{\theta}_g\in\mathrm{R}^T$, formed by the Cartesian product of the base models' outcomes.
In some cases, $f_{\mathcal{D}}^{(t)}(\mathcal{D}_X)$ may be subtly different from the entirety of $\mathcal{Y}$, for instance when algorithms produce unrecognizable output (randomly generated output or external noise). To eliminate the ambiguities, the work presented here relies on the assumption that the outcomes are always in $\mathcal{Y}$.

To learn $\boldsymbol{\theta}_g$, the combiner algorithm uses the outcomes $\left( f_{\mathcal{D}}^{(t)}(\mathcal{D}_X)\right)_{t=1}^T$ that comprise the features of the new $T$-dimensional training set $\Tilde{\mathcal{D}}=(\Tilde{\mathcal{D}}_X, \Tilde{\mathcal{D}}_Y)$, where 

\begin{equation*}
\Tilde{\mathcal{D}}_X=\left\{
(\,f_{\mathcal{D}}^{(1)}(\bm{x}_i), f_{\mathcal{D}}^{(2)}(\bm{x}_i), \ldots, f_{\mathcal{D}}^{(T)}(\bm{x}_i)\,)\right\}_{i=1}^m,
\end{equation*}

\begin{equation*}
\Tilde{\mathcal{D}}_Y = \mathcal{D}_Y.
\end{equation*}
Stacking learns $T+1$ mappings; it learns $f_{\mathcal{D}}^{(t)}:\mathcal{X}\rightarrow\mathcal{Y},t=1,2,\ldots,T$ on the first level using $\mathcal{D}$ as the training set, and then, using these outcomes, it learns one mapping $g:\Tilde{\mathcal{D}}\rightarrow\mathcal{Y}$ on the second level, such that $g\circ f:\,\mathcal{X}\xrightarrow{\Tilde{\mathcal{D}}}\mathcal{Y}$.
Now, the \textit{combiner} learns the parameters $\boldsymbol{\theta}_g$ by minimizing a loss function $L$,

\begin{equation}
\boldsymbol{\theta}_g^* = \argmin_{\boldsymbol{\theta}} L(\boldsymbol{\theta};\, \Tilde{\mathcal{D}}).
\label{eq:stacking}
\end{equation}
The equation of the stacking model $F$ is given by
\begin{equation}
F_{\mathcal{D}}(\bm{x}) = g_{\Tilde{\mathcal{D}}}((f_{\mathcal{D}}^{(1)}(\bm{x}),f_{\mathcal{D}}^{(2)}(\bm{x}),\ldots,f_{\mathcal{D}}^{(T)}(\bm{x}));\boldsymbol{\theta}_g),\quad \bm{x}\in\mathcal{X}
\label{eq:stacking_model}
\end{equation}
and can be reduced to
\begin{equation}
F_{\mathcal{D}}(\bm{x}) = (g_{\Tilde{\mathcal{D}}} \circ f_{\mathcal{D}})(\bm{x}),\quad \bm{x}\in\mathcal{X},
\label{eq:stacking_model_simple}
\end{equation}
where $g$ can simply be regarded as the model equation of the combiner.
For instance, if the combiner is a least-squares regressor, then it minimizes the mean squared error $L(\boldsymbol{\theta}, \Tilde{\mathcal{D}}) = \sum_{z_i\in\mathcal{D}} \left((g \circ f_{\mathcal{D}})(\bm{x}_i)-y_i\right)^2$; classifiers are stacked in the same way, only this time, the mean squared error is replaced by cross-entropy loss. 

Fundamentally, stacking has a conceptual connection to ensemble learning; the parameters $\boldsymbol{\theta}_g^*$---depending on their nature---are simply mixture weights used to combine the $T$ predictors, which is particularly the case for linear combiners. For example, linear or logistic regression learn the weights of predictors, i.e., learn latent coefficients of their linear combination that minimizes the mean squared error or the cross-entropy loss, respectively. The former is nothing but a weighted average of regressors (see Example~\ref{ex:1} below), while logistic regression, to which the combiner establishes weighted majority voting, chooses the weights that minimize the cross-entropy loss and thereon does not treat ensemble members equally.

%
%

\begin{example}{\textbf{(Weighted bagging equals bag-stacking)}}
\label{ex:1}

\noindent
Suppose we are given a bagging ensemble of size $T$. We want to transform it into a weighted ensemble so that each member has a different importance in the final vote (aggregation). To achieve this, a real number $\theta$, called a weight, is associated with each base model. The weights act as measures of relative importance among the $T$ members. We call this \textit{weighted bagging}, and extending Equation~\eqref{eq:agg_bagging} to lend itself to this model, we get

\begin{equation}
f_a(\bm{x};\,\theta) = \mathbf{E}_{\mathcal{D}}[\theta_{\mathcal{D}}f_{\mathcal{D}}(\bm{x})].
\end{equation}

\noindent
To estimate $f_a(\bm{x}; \theta)$, we use $T$ replicates of the training set $\mathcal{D}$, drawn using bootstrap sampling. We assume that the sampling is guided by a random parameter $\bm{r}\in\mathbb{R}^m$ that stores the sampling probabilities of the examples in $\mathcal{D}$. In practice, all elements of $\bm{r}$ are equal to $1/m$. The bootstrap samples are thus $\mathcal{D}(\bm{r}_t), t=1,2,\ldots,T$.

\noindent
Let $\boldsymbol{\theta}=[\theta_{1}\,\theta_{2} \ldots \,\theta_{T}]^\top$ be a latent parameter vector in $\mathbb{R}^T$ that stores the weight of each base model $f_{\mathcal{D}(\bm{r}_t)}$. The objective here is to find $\boldsymbol{\theta}$ that yield the optimal combination of the $T$ base models. The model equation of weighted bagging therefore becomes


\begin{equation*}
F_{\mathcal{D},\bm{r},\boldsymbol{\theta}} (\bm{x}) = 
  \sum_{t=1}^T \theta_{t}f_{\mathcal{D}(\bm{r}_t)}(\bm{x}).
\end{equation*}

\noindent
This equation is applicable to both regression and classification settings. In a classification setting, this is called weighted majority voting. For binary $\{-1, +1\}$ classification tasks, the prediction $\hat{y}$ for any $\bm{x}$ is $\hat{y}=\text{sign}\left[F_{\mathcal{D}, \bm{r}, \boldsymbol{\theta}}(\bm{x})\right]$.

\noindent
To learn $\boldsymbol{\theta}$ after learning $f_{\mathcal{D}(\bm{r}_1)}, f_{\mathcal{D}(\bm{r}_2)}, \ldots, f_{\mathcal{D}(\bm{r}_T)}$, we minimize either a squared loss function in regression,

\begin{equation}
\boldsymbol{\theta}^* = \argmin_{\boldsymbol{\theta}}\sum_{i=1}^m \left(y_i - F_{\mathcal{D}, \boldsymbol{\theta}, \bm{r}}(\bm{x}_i)\right)^2 = \argmin_{\boldsymbol{\theta}}\sum_{i=1}^m \left(y_i - \sum_{t=1}^T f_{\mathcal{D}(\bm{r}_t)}(\bm{x}_i)\right)^2
\label{eq:wb_reg}
\end{equation}

\noindent
or cross-entropy loss in classification, as follows:

\begin{equation}
\boldsymbol{\theta}^* = \argmin_{\boldsymbol{\theta}}-\mathcal{L}(\boldsymbol{\theta};\,\mathcal{D}, F_{\mathcal{D},\bm{r}})=\argmin_{\boldsymbol{\theta}}-\sum_{k=1}^K\sum_{\substack{i=1 \\ y_i=\mathcal{C}_k}}^m \log p(\mathcal{C}_k \, | \, \bm{x}_i,F_{\mathcal{D},\bm{r}}; \boldsymbol{\theta}).
\label{eq:wb_cls}
\end{equation}

\noindent
Setting the partial derivatives with respect to each coordinate in $\boldsymbol{\theta}$ in Equation~\eqref{eq:wb_reg} yields a system of $T$ linear equations with $T$ unknowns,
which can be solved using any relevant method. Equation~\eqref{eq:wb_cls}, by contrast, can be solved using gradient descent or other optimization methods.

\noindent
The takeaway from this example is that Equations~\eqref{eq:wb_reg}~and~\eqref{eq:wb_cls} are in fact the loss functions of the combiner in bag-stacking (linear regression and logistic regression, respectively), hence the relationship of weighted bagging to stacking.

\end{example}

\section{Related work: stability of learning algorithms}
\label{sec:2}

The first conceptions of stability of learning algorithms date back to a few decades ago when machine learning was a fitting part of statistics more than it is today.  
As a general quantitative theory, some of the early notions of \textit{hypothesis} stability stem from~\citep{wolpert1992stacked}.
Being relatively intelligible to a broader audience, the term itself describes the \textit{sensitivity of learning algorithms} as a quantifiable measure of sensitivity to changes in the training set. In the last decade, this notion of stability has been used to derive upper bounds on the generalization error of deterministic and randomized learning algorithms in Elisseeff and Bousquet's work~\citep{elisseeff1,elisseeff_randomized}. Apart from upper bounds based on the Vapnik-Chervonenkis (VC) dimension~\citep{vapnik1999overview} or Probably Approximately Correct (PAC) learnability theory~\citep{valiant1984theory}, stability-based bounds are easily controllable and serve as a powerful tool for designing new learning algorithms.
 
An important virtue of stability-based bounds is their simplicity from a mathematical standpoint. Originally not as tight as the PAC-Bayes bounds~\citep{ambroladze2007tighter}, considered the tightest, they can be optimized or consolidated, i.e., be made significantly tighter in different ways, not the least of which is collaborative learning~\citep{arsov2017generating,pavlovski2018generalization} that has attained significant generalization improvement.

\subsection{Stability and generalization: basic principles for deterministic and randomized learning algorithms}
\label{sec:2.1}

A learning algorithm is stable when it meets a particular stability criterion.
The easiest way to define the stability of a learning algorithm is to start at the goal of establishing an upper bound on its generalization error: we want this bound to be tight when the algorithm meets the stability criterion. More restrictive stability criterions lead to tighter upper bounds~\citep{elisseeff1}. The randomness in supervised learning comes from the sampling of the training set, and stability is thus considered with respect to changes in the training set such as the removal or replacement of a training example~\citep{elisseeff1}. This definition of stability helps establish upper bounds on the generalization error.
Other ways of defining the stability of a learning algorithm and establishing upper bounds include the VC dimension of its search space~\citep{kearns1999algorithmic} and PAC or PAC-Bayes bounds~\citep{ambroladze2007tighter,dasgupta2002pac}.

The rest of this section provides the basics of hypothesis stability, such as notation, variations, and stability for deterministic and randomized algorithms as well as a definition of the generalization error and its upper bound.
The definitions hereinafter apply to both deterministic and randomized learning algorithms, unless explicitly stated otherwise.






\begin{definition}{\textbf{(Loss functions)}}

\noindent
\textit{Let $\ell(y', y) \in \mathbb{R}_0^+$ be a loss function that measures the loss of $y'=f(\bm{x})$ on some $z=(\bm{x},y) \in \mathcal{Z}$. For brevity, we denote this by $\ell(f, z)$.}
\textit{
There are three kinds of loss functions (see Fig.~\ref{fig:loss_functions}):}

\begin{enumerate}
\item \textit{Squared loss} $\ell_{sq}(f,z)\in\mathbb{R},$ $$\ell_{sq}(f,z)=(y-f(\bm{x}))^2.$$

\item \textit{Classification loss} $\ell(f,z)\in\{0,1\},$ $$\ell(f,z)=I(f(\bm{x})\neq y),$$ \textit{where $I(C)$ is an indicator function equal to 1 when $C$ is true, and equal to 0 otherwise.}
\\
\item \textit{The $\gamma$-loss} $\ell_{\gamma}\in\left[0,1\right],$
$$\ell_{\gamma}(f, z) =\begin{cases}
1, & \textrm{ if } yf(\bm{x}) < 0,\\
1 - \displaystyle \frac{yf(\bm{x})}{\gamma}, & \textrm{ if }  0\leq yf(\bm{x})\leq \gamma,\\
0, & \textrm{ otherwise}.
\end{cases}$$ 

\textit{Here, $\ell_{\gamma}(f,z)$ takes into account the margin $m(f,z)$ that increases the loss as it gets closer to zero, which means it favors correct and confident predictions, the intensity being controlled by $\gamma$.}
\end{enumerate}

\begin{figure}[!ht]
\label{fig:loss_functions}
\centering
\includegraphics[width=\textwidth]{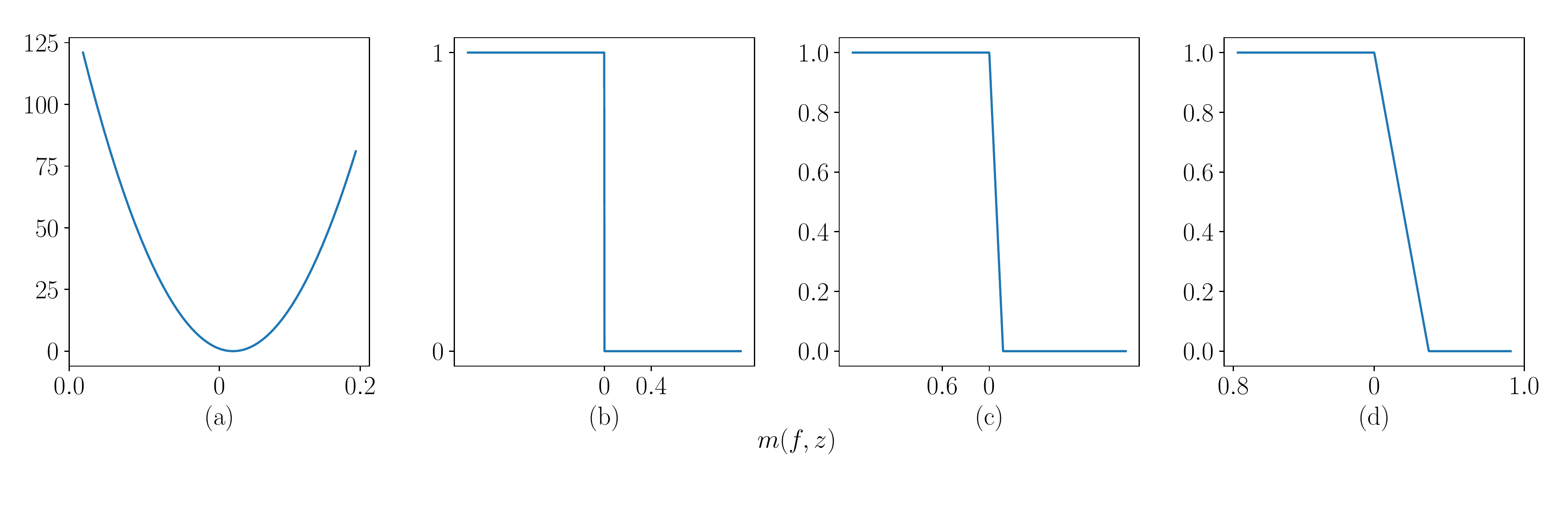}
\caption{Loss functions: the squared loss (a), the classification loss (b), the $\gamma$-loss for $\gamma=1$ and $\gamma=2$, respectively (c and d). Each loss function, barring the squared loss, is confined to $\left[0,1\right]$.}
\end{figure}
\label{def:loss_functions}
\end{definition}

\noindent
Next, we lay out the definitions of the true error of a learning algorithm, also called the generalization error, different notions of stability, and the corresponding upper bounds on the generalization error. To begin with, let $f_{\mathcal{D}}(\bm{x})$ be the outcome of a learning algorithm trained on $\mathcal{D}$.

\begin{definition}{\textbf{(Generalization error)}}

\noindent
\textit{Let $\ell$ be a loss function. The true, i.e., generalization error of a learning algorithm that trains a model $f$ using $\mathcal{D}$ is expressed as the expected loss of $f$ for all $z\in\mathcal{Z}$, that is,}

\begin{equation}
R_{gen}(f_{\mathcal{D}})=\mathbf{E}_z[\ell(f_{\mathcal{D}}, z)].
\label{eq:gen_err}
\end{equation}

\label{def:gen_err}
\end{definition}

\noindent
The simplest estimator of $R(f,\mathcal{D})$ is the empirical error observed on the sample $\mathcal{D}$, also known as the training error,

\begin{equation}
R_{emp}(f,\mathcal{D})=\frac{1}{m}\sum_{i=1}^m \ell(f_{\mathcal{D}}, z_i).
\label{eq:emp_err}
\end{equation}

\noindent
In addition, the leave-one-out error is given by

\begin{equation}
R_{loo}(f, \mathcal{D}) = \frac{1}{m} \sum_{i=1}^m \ell(f_{\mathcal{D}^{\backslash i}},z_i).
\label{eq:loo_err}
\end{equation}

\begin{definition}{\textbf{(Hypothesis stability)} \citep[Def. 1]{elisseeff_randomized}}

\noindent
\textit{A learning algorithm has hypothesis stability $\beta$ with respect to a loss function $\ell$ if the following holds:}

\begin{equation}
\forall i \in \{1,2,\ldots,m\},\quad \mathbf{E}_{\mathcal{D},z}\left[|\ell(f_{\mathcal{D}},z) - \ell(f_{\mathcal{D}^{\backslash i}},z)|\right] \leq \beta.
\label{eq:hyp_stability}
\end{equation}

\label{def:hyp_stability}
\end{definition}

\noindent
With this definition of stability, the expected loss difference is measured on a fixed example $z\in\mathcal{Z}$ while one example at a time is removed from the training set $\mathcal{D}$. This strategy provides the means to derive an upper bound on the generalization error based on the leave-one-out error $R_{loo}(f, \mathcal{D})$ of the algorithm.

\begin{theorem}{\textup{\textbf{(Hypothesis stability generalization error bound)} \citep[Thm. 2]{elisseeff_randomized}}}

\noindent
Let $f_{\mathcal{D}}$ be the outcome of a learning algorithm with hypothesis stability $\beta$ with respect to a loss functions $\ell$ such that $0 \leq \ell(f, z) \leq M$. Then, with probability $1 - \delta$ over the random draw of the training set $\mathcal{D}$	,

\begin{equation}
R_{gen}(f_{\mathcal{D}}) \leq R_{loo}(f_{\mathcal{D}}, \mathcal{D}) + \sqrt{{\delta}^{-1}\frac{M^2 + 6Mm\beta}{2m}}.
\label{eq:hyothesis_stability_bound}
\end{equation}

\label{thm:hypothesis_stability_bound}
\end{theorem}

A slightly different notion of stability, called \textit{pointwise} hypothesis stability, measures the loss change on one of the training examples $z_i \in \mathcal{D}$ that is replaced by some $z\in\mathcal{Z}$, which is not originally in $\mathcal{D}$. 

\begin{definition}{\textbf{(Pointwise hypothesis stability)}~\citep[Def. 3]{elisseeff_randomized}}

\noindent
\textit{A learning algorithm has pointwise hypothesis stability $\beta$ with respect to the loss function $\ell$ if the following holds:}

\begin{equation}
\forall i \in \{1,2,\ldots,m\},\quad \mathbf{E}_{\mathcal{D}}\left[|\ell(f_{\mathcal{D}},z_i) - \ell(f_{\mathcal{D}^{\backslash i}\cup z},z_i)|\right] \leq \beta.
\label{eq:point_hyp_stability}
\end{equation}

\label{def:point_hyp_stability}
\end{definition}

\noindent
With pointwise hypothesis stability we get a similar error bound that can now include the empirical (training) error $R_{emp}(f_{\mathcal{D}}, \mathcal{D})$ of the algorithm.

\begin{theorem}{\textup{\textbf{(Pointwise hypothesis stability generalization error bound)}~\citep[Thm. 4]{elisseeff_randomized}}}

\noindent
Let $f_{\mathcal{D}}$ be the outcome of a learning algorithm with pointwise hypothesis stability $\beta$ with respect to a loss function $\ell$ such that $0 \leq \ell(f,z) \leq M$. Then, with probability $1 - \delta$ over the random draw of the training set $\mathcal{D}$,

\begin{equation}
R_{gen}(f_{\mathcal{D}}) \leq R_{emp}(f_{\mathcal{D}}, \mathcal{D}) + \sqrt{\delta^{-1} \frac{M^2 + 12Mm\beta}{2m}}.
\label{eq:pontwise_hypothesis_stability_bound}
\end{equation}

\label{thm:pointwise_hypothesis_stability_bound}
\end{theorem}

An even stronger notion of stability, called uniform stability, provides tighter bounds on the generalization error.

\begin{definition}{\textbf{(Uniform stability)} \citep[Def.5]{elisseeff_randomized}}

\noindent
\textit{An algorithm has uniform stability $\beta$ with respect to the loss function $\ell$ if the following holds}

\begin{equation}
\forall \mathcal{D} \in \mathcal{Z}^m,\, \forall i\in\{1,\ldots,m\},\quad \| \ell(f_{\mathcal{D}},\, .) - \ell(f_{\mathcal{D}^{\backslash i}},\, .)\|_{\infty} \leq \beta
\label{eq:uniform_stability}
\end{equation}

\label{def:uniform_stability}
\end{definition}

\noindent
Uniform stability is an upper bound on hypothesis and pointwise hypothesis stability~\citep{elisseeff1}.

\begin{theorem}{\textup{\textbf{(Uniform stability generalization error bound)} \citep[Thm. 6]{elisseeff_randomized}}}

\noindent
Let $f_{\mathcal{D}}$ be the outcome of a learning algorithm with uniform stability $\beta$ with respect to a loss function $\ell$ such that $0 \leq \ell(f,z) \leq M$, for all $z \in \mathcal{Z}$ and all sets $\mathcal{D}$. Then, for any $m \geq 1$ and any $\delta \in (0, 1)$, the following bound holds with probability $1-\delta$ over the random draw of the training set $\mathcal{D}$,

\begin{equation}
R_{gen}(f_{\mathcal{D}}) \leq R_{emp}(f_{\mathcal{D}}, \mathcal{D}) + 2 \beta + (4m\beta + M) \sqrt{\frac{\log (1/\delta)}{2m}}.
\label{eq:uniform_stability_bound}
\end{equation}

\label{thm:uniform_stability_bound}
\end{theorem}

\noindent
In this bound, the dependence is $\sqrt{\log (1/\delta)}$ instead of $1/\delta$, which implies a tighter upper bound on the generalization error.

\noindent
Algorithms whose stability scales as $O(1/m)$ are considered \textit{stable}~\citep{elisseeff1}.

When an algorithm is randomized using random parameter $\bm{r}$, we then have a randomized $f_{\mathcal{D},\bm{r}}$ and consequently, random hypothesis, pointwise hypothesis, and uniform stability definitions. When $\bm{r}$ is fixed, they are equal to Definitions~\ref{def:hyp_stability}~and~\ref{def:point_hyp_stability}. The randomized definitions of stability and the resulting upper bounds on the generalization error are strikingly similar to the deterministic ones.

\begin{definition}{\textbf{(Uniform stability of randomized algorithms)}~\citep[Def. 13]{elisseeff_randomized}}

\noindent
\textit{A randomized learning algorithm has uniform stability $\beta$ with respect to the loss function $\ell$ if for every $i=1,\ldots,m$,}

\begin{equation}
\sup_{\mathcal{D},\bm{r}} \left| \mathbf{E}_{\bm{r}}\left[\ell(f_{\mathcal{D},\bm{r}}, z)\right] - \mathbf{E}_{\bm{r}}\left[\ell(f_{\mathcal{D}^{\backslash i},\bm{r}},z)\right|\right] \leq \beta.
\label{eq:random_uniform_stability}
\end{equation}

\label{def:random_uniform_stability}
\end{definition}



The empirical estimate $R_{emp}(f,\mathcal{D})$ can be used to construct a guess for the generalization error concealed behind the unknown distribution $P$ of $\mathcal{Z}$. Stability of learning algorithms can then be leveraged to construct an upper bound on the generalization error.

\subsection{Stability and generalization error upper bounds for bagging and subbagging}

This section focuses on bagging and subbagging. The notation conforms to the one used in~\cite{elisseeff_randomized}. 
Let $T$ be the number of bootstrap samples $\mathcal{D}(\bm{r}_1), \ldots , \mathcal{D}_{\bm{r}_T}$, where $\mathcal{D}(\bm{r}_t)$ denotes the $t$-th bootstrap set. The random parameter $\bm{r}_t\in \mathcal{R}=\{1,\ldots,m\}^m$ are instances of a random variable corresponding to random sampling \textit{with} replacement of $m$ elements from the training set $\mathcal{D}$, and such random variables have a multinomial distribution with parameters $\left(\frac{1}{m},\ldots,\frac{1}{m}\right)$, which means each example in $\mathcal{D}$ has an equal probability to be sampled~\citep{elisseeff_randomized}. For simplicity, the fact that the base algorithm can also be randomized is omitted. The bagging model can thus be written as 

\begin{equation*}
F_{\mathcal{D},\bm{r}}=\frac{1}{T} \sum_{t=1}^T f_{\mathcal{D}(\bm{r}_t)}.
\end{equation*}

\begin{proposition}{\textup{\textbf{(Random hypothesis stability of bagging for regression)} \citep[Prop. 4.1]{elisseeff_randomized}.}}

\noindent
Let the loss $\ell$ be $B$-Lipschitzian with respect to its first variable and let $F_{\mathcal{D},\bm{r}}$ be the outcome of a bagging algorithm whose base algorithm $f_{\mathcal{D}(\bm{r}_t)}$ has hypothesis (respectfully pointwise hypothesis) stability $\gamma_m$ with respect to the $\ell_1$ loss function. Then, the random hypothesis (respectfully pointwise hypothesis) stability $\beta_m$ of $F_{\mathcal{D},\bm{r}}$ with respect to $\ell$ is bounded by

\begin{equation}
\beta_m \leq B\sum_{k=1}^m \frac{k \gamma_k}{m}\mathbb{P}_{\bm{r}}\left[d(\bm{r})=k\right],
\label{eq:bagging_stability_reg}
\end{equation}

\noindent
where $d(\bm{r}),\bm{r}\in \mathcal{R}$ is the number of distinct examples in one bootstrap iteration.
\label{thm:bagging_stability_reg}
\end{proposition}

\begin{proposition}{\textup{\textbf{(Random hypothesis stability of bagging for classification)} \citep[Prop. 4.2]{elisseeff_randomized}.}}

\noindent
Let $F_{\mathcal{D},\bm{r}}$ be the outcome of a bagging algorithm whose base algorithm $f_{\mathcal{D}(\bm{r}_t)}$ has hypothesis (respectfully pointwise hypothesis) stability $\gamma_m$ with respect to the classification loss function. Then, the random hypothesis (respectfully pointwise hypothesis) stability $\beta_m$ of $F_{\mathcal{D},\bm{r}}$ with respect to the $\ell_1$ loss is bounded by

\begin{equation}
\beta_m \leq 2\sum_{k=1}^m \frac{k \gamma_k}{m}\mathbb{P}_{\bm{r}}\left[d(\bm{r})=k\right].
\label{eq:bagging_stability_cls}
\end{equation}

\label{thm:bagging_stability_cls}
\end{proposition}





The upper bounds for subbagging are less complicated since the sampling is \textit{without} replacement, meaning that the examples in each training set $\mathcal{D}(\bm{r}_t)$ are distinct.

\begin{proposition}{\textup{\textbf{(Stability of subbagging for regression)} \citep[Prop. 4.4]{elisseeff_randomized}.}}
Assume that the loss $\ell$ is $B$-Lipschitzian with respect to its first variable. Let $F_{\mathcal{D},\bm{r}}$ be the outcome of a subbagging algorithm whose base algorithm has uniform (respectfully hypothesis or pointwise hypothesis) stability $\gamma_p$ with respect to the $\ell_1$ loss function, and subbagging is done by sampling $p\leq m$ points without replacement. Then, the random uniform (respectfully hypothesis or pointwise hypothesis) stability $\beta_m$ of $F_{\mathcal{D},\bm{r}}$ with respect to $\ell$ is bounded by

\begin{equation}
\beta_m \leq B\gamma_p\frac{p}{m}.
\label{eq:subbagging_stability_reg}
\end{equation}

\label{thm:subbagging_stability_reg}
\end{proposition}

\begin{proposition}{\textup{\textbf{(Stability of subbagging for classification)} \citep[Prop. 4.5]{elisseeff_randomized}.}}
Let $F_{\mathcal{D},\bm{r}}$ be the outcome of a subbagging algorithm whose base algorithm has hypothesis (respectfully pointwise hypothesis) stability $\gamma_p$ with respect to the classification loss function, and subbagging is done by subsampling $p \leq m$ examples from $\mathcal{D}$ without replacement. Then, the random hypothesis (respectfully pointwise hypothesis) stability $\beta_m$ of $F_{\mathcal{D},\bm{r}}$ with respect to $\ell_1$ loss is bounded by

\begin{equation}
\beta_m \leq 2\gamma_p \frac{p}{m}.
\label{eq:subbagging_stability_cls}
\end{equation}

\label{thm:subbagging_stability_cls}
\end{proposition}

\noindent
The propositions above can be used to derive an upper bound on the generalization errors of bagging and subbagging. The following theorems give the upper bounds for subbagging, the latter being tighter, where the dependence is $\sqrt{\log (2/\delta)}$ rather than $\sqrt{1/\delta}$. Moreover, the same bounds hold for bagging, where $\frac{p\gamma_p}{m}$ is replaced by $\sum_{k=1}^m \frac{k\gamma_k}{m}\mathbb{P}_{\bm{r}}[d(\bm{r})=k]$, which is roughly $0.632 \gamma_{0.632m}$ when $m$ is sufficiently large.

\begin{theorem}{\textup{\textbf{(Hypothesis and pointwise hypothesis stability upper bound on the generalization error of subbagging)} \citep[Thm. 16]{elisseeff_randomized}}}

\noindent
Assume that the loss $\ell$ is $B$-lipschitzian with respect to its first variable. Let $F_{\mathcal{D},\bm{r}}$ be the outcome of a subbagging algorithm. Assume subbagging is done with $T$ sets of size $p$ subsampled without replacement from $\mathcal{D}$ and the base learning algorithm has hypothesis stability $\gamma_p$ and pointwise hypothesis stability $\gamma_p'$ with respect to the loss $\ell$. The following bounds hold separately with probability at least $1-\delta$

\begin{equation}
R_{gen}(F_{\mathcal{D},\bm{r}}) \leq R_{loo}(F_{\mathcal{D},\bm{r}}), \mathcal{D}) + \sqrt{\delta^{-1}\frac{2M^2 + 12 MB p \gamma_p}{m}},
\label{eq:hypothesis_pointwise_hypothesis_error_bound_subbagging_loo}
\end{equation}

\begin{equation}
R_{gen}(F_{\mathcal{D},\bm{r}}) \leq R_{emp}(F_{\mathcal{D},\bm{r}}), \mathcal{D}) + \sqrt{\delta^{-1}\frac{2M^2 + 12 MB p \gamma_p'}{m}}.
\label{eq:hypothesis_pointwise_hypothesis_error_bound_subbagging_emp}
\end{equation}

\label{thm:hypothesis_pointwise_hypothesis_error_bound_subbagging}
\end{theorem}






The bounds above are derived for subbagging, but the same bounds hold true for bagging~\citep{elisseeff_randomized} in case $\frac{p\gamma_p}{m}$ is replaced by $\sum_{i=1}^m \frac{i\gamma_i}{m} \mathbb{P}(d(\bm{r})=i)$ which is roughly equal to $0.632\gamma_{0.632m}$, where $d(\bm{r}), \bm{r} \in \mathbb{R}^T$ is the number of distinct sampled points in one bootstrap iteration.

\section{Stability analysis of bagging, stacking and bag-stacking: why and when do they work?}
\label{sec:main}

This section is devoted to providing readers with a thorough view of how stability interacts in bagging, stacking, and in-between. 
Here we emphasize that stability is the appropriate formal apparatus to argue the effectiveness of stacking, more so than other relatively lenient, yet pragmatically acclaimed expositions. To illustrate one, the following statement appears to convey cogent reasoning: \textit{"Students in a school are classified into three groups: bad, average, and excellent. The confusion matrix shows that a K-NN classifier does well only in discerning bad and excellent students; a decision tree classifier, on the other hand, does so with average and excellent. Now, stacking the K-NN and decision tree models together, one might expect to obtain a predictor that accurately discerns all three classes of students."}

While a trial-and-error approach might prove this statement correct, the conventional wisdom used here does not suffice, hence the need to formally investigate the effectiveness of stacking. The upper bound on the generalization error of a learning algorithm becomes tighter as its stability increases, i.e., as the value of the stability measure decreases. Throughout this section, we are going to show that---as opposed to the conventional heterogeneous nature of the stacking ensemble---its homogeneous counterpart, bagging, can instead be used on the first level in the stacking pipeline to further improve stability and, as a consequence, yield a tighter upper bound on the generalization error and hence improve performance.

Bag-stacking can be seen as a simple modification of stacking, where instead of passing the predictions by different kinds of models for the combiner algorithm to learn, we use several models of the same kind, but learned using different training sets constructed via bootstrap sampling. Being perhaps the simplest way to achieve this, the predictor aggregation step of bagging is omitted and subsequently replaced by the outcome of a combiner algorithm. It lends a heterogeneity property to stacking and can be used to overcome various shortcomings of stacking (bootstrap sampling generates different training sets and thus reduces the variance of the stacking model) and bagging (stacking the bagged models introduces weights of the models unlike bagging that treats all base models equally). The results in~\citep{bag_stacking} show that bag-stacking and its variation, dag-stacking, usually outperform both bagging and stacking.

Bag-stacking, however, has never attained the whopping success of bagging and stacking in the research community as have had bagging and stacking, though it is clear that it integrates their traits. The work on bag-stacking is comprised of a scant paper~\citep{bag_stacking} that compares its error rate to those of bagging, stacking, and a special case referred to as \textit{dag-stacking}, where subsets of unique points are used in place of bootstrap samples. The paper is, however, wanting in theoretical analysis as well as a formal explanation of bag-stacking and dag-stacking's effectiveness.

\subsection{Stability of Stacking}
\label{sec:stacking_stability}

In this part, we analyze the hypothesis stability of stacking according to Definition~\ref{def:hyp_stability}. Here, we use the classification loss given in Definition~\ref{def:loss_functions}. The smooth $\ell_1$ loss is also applicable because $\ell_1(y',y)\leq\ell(y',y)$ for all $y, y'\in\mathcal{Y}$.

To analyze interaction of hypothesis stability among the constituent models in stacking, we first look at the expected absolute loss difference with respect to $z\in\mathcal{Z}$, i.e., $\mathbf{E}_{z}[|\ell(f_{\mathcal{D}}, z) - \ell(f_{\mathcal{D}^{\backslash i}}, z)|]$ and then take the expectation with respect to $\mathcal{D}$ in order to get $\mathbf{E}_{\mathcal{D},z}[|\ell(f_{\mathcal{D}}, z) - \ell(f_{\mathcal{D}^{\backslash i}}, z)|]$. The goal is to bound $\mathbf{E}_{z}[|\ell(f_{\mathcal{D}}, z) - \ell(f_{\mathcal{D}^{\backslash i}}, z)|]$ from above and the changes in the outcome $f_{\mathcal{D}}(\bm{x})$ when $z_i$ is removed from $\mathcal{D}$. For instance, in the $k$-NN algorithm, the loss difference is bounded by the probability of the set of examples $v_i$ such that the closest point from the training set to any point in $v_i$ is $z_i$, or $v_i=\{z'\,|\,\text{dist}(z',z_i)\leq \text{dist}(z', z_j) \land j\neq i\}$; the loss difference depends on $\mathbb{P}(v_i)$ and thus for $z\in\mathcal{Z}$, it holds that $\mathbf{E}_{z}[|\ell(f_{\mathcal{D}}, z) - \ell(f_{\mathcal{D}^{\backslash i}}, z)|]\leq \mathbb{P}(v_i)$~\citep[Example 1]{elisseeff_randomized}. We apply the same logic here, noting that when dealing with ensembles, one needs to take into account the stability of the base learning algorithm(s).
To analyze the hypothesis stability of stacking, it is important to stress that

\begin{itemize}
    \item[$(i)$] the base learning algorithms are applied independently of one another, and\\
    \item[$(ii)$]the combiner learning algorithm is applied independently of the base learning algorithms.
\end{itemize}

\noindent
Consequently, the stability of each base algorithm is independent of the rest and the stability of the combiner algorithm is also independent of that of the base algorithms. 

\noindent
Next, we continue by formally expressing the statements above: $(i)$ for the outcomes of each pair of base models $f_{\mathcal{D}}^{(t)}(\bm{x})$ and $f_{\mathcal{D}}^{(s)}(\bm{x})$, it holds that 

\begin{equation}
    \mathbf{E}_z\left[|\ell(f_{\mathcal{D}}^{(t)}, z) - \ell(f_{\mathcal{D}^{\backslash i}}^{(t)}, z)|\right]\mathbf{E}_z\left[|\ell(f_{\mathcal{D}}^{(s)}, z) - \ell(f_{\mathcal{D}^{\backslash i}}^{(s)}, z)|\right]\leq\mathbb{P}_{f_{\mathcal{D}}^{(t)}}\mathbb{P}_{f_{\mathcal{D}}^{(s)}},\quad t\neq s
    \label{eq:stacking_eq_1}
\end{equation}

\noindent
which follows from multiplying the inequalities $\mathbf{E}_z[|\ell(f_{\mathcal{D}}^{(j)}, z) - \ell(f_{\mathcal{D}^{\backslash i}}^{(j)}, z)|]\leq\mathbb{P}_{f_{\mathcal{D}}^{(j)}}$ and $\mathbf{E}_z[|\ell(f_{\mathcal{D}}^{(k)}, z) - \ell(f_{\mathcal{D}^{\backslash i}}^{(k)}, z)|]\leq\mathbb{P}_{f_{\mathcal{D}}^{(k)}}$,
whereas $(ii)$ for the outcome $g_{\Tilde{\mathcal{D}}}(\bm{x})$ of the combiner algorithm and the outcome $f_{\mathcal{D}}(\bm{x})$ of a base algorithm, it holds that 

\begin{equation}
    \mathbf{E}_z\left[|\ell(f_{\mathcal{D}}, z) - \ell(f_{\mathcal{D}^{\backslash i}}, z)|\right]\mathbf{E}_{\Tilde{z}}\left[|\ell(g_{\Tilde{\mathcal{D}}}, z) - \ell(f_{\Tilde{\mathcal{D}}^{\backslash i}}, z)|\right]\leq\mathbb{P}_{f_{\mathcal{D}}}\mathbb{P}_{g_{\Tilde{\mathcal{D}}}},
    \label{eq:stacking_eq_2}
\end{equation}
where $\Tilde{D} = (f_{\mathcal{D}}^{(t)}(\mathcal{D}_X))_{t=1}^T$.

\noindent
Combining Equations~\eqref{eq:stacking_eq_1}~and~\eqref{eq:stacking_eq_2} into a single equation for a stacking model yields the following bound on the expected absolute loss difference with respect to $\Tilde{z}\in\mathcal{Z}$ such that for $z\in\mathcal{Z}$, $\Tilde{z}$ has the same output. i.e., $\Tilde{y}=y$:

\begin{equation}
    \mathbf{E}_{\Tilde{z}}\left[|\ell(g_{\Tilde{\mathcal{D}}}, \Tilde{z}) - \ell(g_{\Tilde{\mathcal{D}}^{\backslash i}}, \Tilde{z})|\right] \leq \mathbb{P}_{g_{\Tilde{\mathcal{D}}}}\prod_{t=1}^T \mathbb{P}_{f_{\mathcal{D}}^{(t)}}.
    \label{eq:stacking_hypstb_1}
\end{equation}

\noindent
Finally, taking $\mathbf{E}_{\mathcal{D}}$ of both sides of Equation~\eqref{eq:stacking_hypstb_1} to get hypothesis stability, we get

\begin{equation}
    \mathbf{E}_{\mathcal{D},\Tilde{z}}\left[|\ell(g_{\Tilde{\mathcal{D}}}, \Tilde{z}) - \ell(g_{\Tilde{\mathcal{D}}^{\backslash i}}, \Tilde{z})|\right] \leq \mathbf{E}_{\mathcal{D}}\left[ \mathbb{P}_{g_{\Tilde{\mathcal{D}}}}\prod_{t=1}^T \mathbb{P}_{f_{\mathcal{D}}^{(t)}}\right]=\mathbf{E}_{\Tilde{\mathcal{D}}}[\mathbb{P}_{g_{\Tilde{\mathcal{D}}}}]\prod_{t=1}^T \mathbf{E}_{\mathcal{D}}[\mathbb{P}_{f_{\mathcal{D}}^{(t)}}].
    \label{eq:stacking_hypstb_2}
\end{equation}

\noindent
Note that taking $\mathbf{E}_{\mathcal{D}}$ has the same effect as taking $\mathbf{E}_{\Tilde{\mathcal{D}}}$. The expectations $\mathbf{E}_{\mathcal{D}}[\mathbb{P}_{g_{\hat{\mathcal{D}}}}]$ and $\mathbf{E}_{\mathcal{D}}[\mathbb{P}_{f_{\mathcal{D}}}^{(t)}]$ are essentially the hypothesis stability expressions $\beta(g_{\Tilde{\mathcal{D}}})=\mathbf{E}_{\mathcal{D}}[\mathbb{P}_{f_{\Tilde{\mathcal{D}}}^{(t)}}]$ and $\beta(f^{(t)})=\mathbf{E}_{\mathcal{D}}[\mathbb{P}_{f_{\mathcal{D}}^{(t)}}]$, as given in Definition~\ref{def:hyp_stability}. Finally, the hypothesis stability of stacking is

\begin{equation}
    \mathbf{E}_{\mathcal{D},\Tilde{z}}\left[|\ell(g_{\Tilde{\mathcal{D}}}, \Tilde{z}) - \ell(g_{\Tilde{\mathcal{D}}^{\backslash i}}, \Tilde{z})|\right] \leq \beta(g_{\Tilde{\mathcal{D}}}) \prod_{t=1}^T \beta(f_{\mathcal{D}}^{(t)}).
    \label{eq:stacking_hypothesis_stability}
\end{equation}\\
The rightmost equality in Equation~\eqref{eq:stacking_hypstb_2} follows from statements $(i)$ and $(ii)$. In other words, the hypothesis stability of stacking is the product of the hypothesis stabilities of all the base models and the combiner. The independence between the base and combiner algorithms eases the computations. 
Equation~\eqref{eq:stacking_hypothesis_stability} also shows that increasing the number of base models improves the stability of stacking.

For example, let there be a stacking ensemble in which a ridge regression classifier with a penalty $\lambda$ acts as the combiner and there are three base $k$-NN classifiers with $k_1, k_2$, and $k_3$, with hypothesis stability $1/\lambda m$, $k_1/m$, $k_2/m$, and $k_3/m$, respectively.
The hypothesis stability of this stacking ensemble is 

\begin{equation}
    \mathbf{E}_{\Tilde{\mathcal{D}}, \Tilde{z}}\left[ |\ell(g_{\Tilde{\mathcal{D}}}, \Tilde{z}) - \ell(g_{\Tilde{\mathcal{D}}^{\backslash i}}, \Tilde{z})|\right] \leq \frac{k_1k_2k_3}{\lambda m^4}.
    \label{eq:hypothesis_stability_of_stacking_eg}
\end{equation}



\subsection{Stability of dag-stacking and bag-stacking}
\label{sec:bagstacking_stability}

In bag-stacking, the only change is that, now, the base models are trained on bootstrap samples drawn from $\mathcal{D}$ instead of using $\mathcal{D}$ to train them. In addition, the bootstrap samples still allow one to use different base learning algorithms. When stacking is combined with bagging, it is easy to see that the merger of the two is a weighted bagging ensemble. We continue by analyzing the hypothesis stability of bag-stacking and dag-stacking.

\subsubsection{Hypothesis stability of bag-stacking: why bootstrap sampling improves stacking?}

In this part, we describe how bootstrap sampling improves stacking. At the same time, Equation~\eqref{eq:bag_stacking_2} gives the hypothesis stability of bag-stacking.

Recall that for any $z=(\bm{x},y) \in \mathcal{Z}$, $\Tilde{z}=((f_{\mathcal{D}}^{(t)})_{t=1}^T, y)\in\mathcal{Z}$. Since the second coordinate of $z$ and $\Tilde{z}$ is the same, we can say that the expected value with respect to $z$, i.e., $\mathbf{E}_{z}$ is proportional to the one with respect to $\Tilde{z}$, $\mathbf{E}_{\Tilde{z}}$.

Let $N_i$ be the number of bootstrap samples in which the training example $z_i$ appears. The probability of any training example appearing in a bootstrap sample is $0.632/m$. Thus, $N_i$ follows a Binomial distribution with $p=0.632/m$ and $n=T$. In ensemble learning, we are interested whether $N_i > s$, for $1\leq s \leq T$, and we thus have a sum of Binomials, given that $N_i\sim B(0.632/m, T)$

\begin{equation}
    \mathbb{P}(N_i > s) = \sum_{k=s+1}^T \binom{T}{k} \left(\frac{0.632}{m}\right)^k\left(1-\frac{0.632}{m}\right)^{T - k}.
    \label{eq:binomial}
\end{equation}

\noindent
For bag-stacking, the expected absolute loss difference depends on whether $z_i$ appears in more than half of the bootstrap samples, i.e., on $\mathbb{P}(N_i > T/2)$. Therefore, the hypothesis stability of stacking with bootstrap sampling is

\begin{equation}
    \mathbf{E}_{\Tilde{D},\Tilde{z}}\left[|\ell(g_{\Tilde{D}}, \Tilde{z}) - \ell(g_{\Tilde{D}^{\backslash i}}, \Tilde{z})|\right] \leq \mathbb{P}(N_i > \frac{T}{2})\mathbf{E}_{\Tilde{\mathcal{D}}}[\mathbb{P}_{g_{\Tilde{\mathcal{D}}}}]\prod_{t=1}^T\mathbf{E}_{\mathcal{D}}[\mathbb{P}_{f_{\mathcal{D}}^{(t)}}],
    \label{eq:bag_stacking_1}
\end{equation}
i.e.,
\begin{equation}
    \mathbf{E}_{\Tilde{D},\Tilde{z}}\left[|\ell(g_{\Tilde{D}}, \Tilde{z}) - \ell(g_{\Tilde{D}^{\backslash i}}, \Tilde{z})|\right] \leq \mathbb{P}(N_i > \frac{T}{2})\beta(g_{\Tilde{\mathcal{D}}})\prod_{t=1}^T\beta(f_{\mathcal{D}}^{(t)}),
    \label{eq:bag_stacking_2}
\end{equation}
where $\mathbb{P}(N_i > T/2)=\sum_{k=\lfloor{T/2}\rfloor +1}^T \binom{T}{k}(0.632/m)^k(1-0.632/m)^{T-k}$.\\
In classical stacking, without bootstrap sampling, $N_i=T$, which means that $\mathbb{P}(N_i>T/2)=1$ and the hypothesis stability reduces to Equation~\eqref{eq:stacking_hypothesis_stability}. 
Therefore, we can conclude that bootstrap sampling improves the stability of stacking by a factor proportional to the order of $\mathbb{P}(N_i > T/2)$. 
The whole method is known as bag-stacking.

Following the example given in Section~\ref{sec:stacking_stability}, the hypothesis stability of a bag-stacking ensemble, arranged in the same way as the stacking ensemble given there, is 

\begin{flalign}
    \mathbf{E}_{\Tilde{\mathcal{D}}, \Tilde{z}}\left[|\ell(g_{\Tilde{\mathcal{D}}}, \Tilde{z}) - \ell(g_{\Tilde{\mathcal{D}}^{\backslash i}}, \Tilde{z})| \right] &\leq \mathbb{P}(N_i > 1)\frac{k_1k_2k_3}{\lambda m^4}\nonumber\\&=\frac{k_1k_2k_3}{\lambda m^4}\sum_{k=0}^1\binom{3}{k}\left(\frac{0.632}{m}\right)^k \left(1-\frac{0.632}{m}\right)^{3-k}
    \label{eq:hypothesis_stability_of_bag_stacking_eg}
\end{flalign}

\subsubsection{Hypothesis stability of dag-stacking: why subsampling improves stacking?}

In this part, we explain why subsampling improves stacking. At the same time, Equation~\eqref{eq:dag_stacking_2} gives the hypothesis stability of dag-stacking.

In dag-stacking, we use subsampling instead of bootstrap sampling on top of stacking. $T$ subsamples of size $p<m$ are drawn from $\mathcal{D}$ without replacement. Thus, the probability that a training example $z_i$ appears in one of the subsamples is $1/p$ instead of $0.632/m$. The resulting hypothesis stability of dag-stacking is the same as in Equation~\eqref{eq:bag_stacking_2}, i.e., we have the same inequality
\begin{equation}
    \mathbf{E}_{\Tilde{D},\Tilde{z}}\left[|\ell(g_{\Tilde{D}}, \Tilde{z}) - \ell(g_{\Tilde{D}^{\backslash i}}, \Tilde{z})|\right] \leq \mathbb{P}(N_i > \frac{T}{2})\beta(g_{\Tilde{\mathcal{D}}})\prod_{t=1}^T\beta(f_{\mathcal{D}}^{(t)}),
    \label{eq:dag_stacking_2}
\end{equation}
except that, this time, $\mathbb{P}(N_i > T/2)=\sum_{k=\lfloor{T/2}\rfloor +1}^T \binom{T}{k}(1/p)^k(1-1/p)^{T-k}$.
Again, we can conclude that subsampling improves the stability of stacking by a factor proportional to the order of $\mathbb{P}(N_i > T/2)$ with respect to $p$.

For example, a dag-stacking ensemble in which the subsamples drawn from $\mathcal{D}$ are of size $p$, has hypothesis stability

\begin{flalign}
    \mathbf{E}_{\Tilde{\mathcal{D}}, \Tilde{z}}\left[|\ell(g_{\Tilde{\mathcal{D}}}, \Tilde{z}) - \ell(g_{\Tilde{\mathcal{D}}^{\backslash i}}, \Tilde{z})| \right] &\leq \mathbb{P}(N_i > 1)\frac{k_1k_2k_3}{\lambda m^4}\nonumber\\&=\frac{k_1k_2k_3}{\lambda m^4}\sum_{k=0}^1\binom{3}{k}\left(\frac{p}{m}\right)^k \left(1-\frac{p}{m}\right)^{3-k}
    \label{eq:hypothesis_stability_of_dag_stacking_eg}
\end{flalign}

\subsection{Why stacking improves bagging and subbagging?}

In this part, we show that using a combiner on top of bagging or subbagging improves the stability by a factor proportional to the hypothesis stability of the combiner learning algorithm. 
This way, the model becomes ``weighted" bagging, or ``weighted" subbagging, because the ensemble members are not treated equally in the final majority vote.
It is necessary to emphasize again that the stability of bagging (or subbagging, respectively) and the stability of the combiner are independent.

According to Equations~\eqref{eq:bagging_stability_cls}~and~\eqref{eq:subbagging_stability_cls} (hypothesis stability of bagging and subbagging where the base algorithms have stabilities $\gamma_m$ and $\gamma_k$, respectively), adding a combiner $g_{\Tilde{\mathcal{D}}}$ trained on $\Tilde{\mathcal{D}}$ yields the following inequalities with respect to a $B$-Lipschitzian loss function $\ell$:

\begin{equation}
    \mathbf{E}_{\Tilde{D}, \Tilde{z}}\left[|\ell(g_{\Tilde{\mathcal{D}}}, \Tilde{z}) - \ell(g_{\Tilde{\mathcal{D}}^{\backslash i}}, \Tilde{z}) |\right]\leq B\beta(g_{\Tilde{\mathcal{D}}})\sum_{k=1}^m \frac{k\gamma_k}{m}\mathbb{P}_{\bm{r}}[d(\bm{r})=k],
    \label{eq:weighted_bagging}
\end{equation}

\begin{equation}
    \mathbf{E}_{\Tilde{D}, \Tilde{z}}\left[|\ell(g_{\Tilde{\mathcal{D}}}, \Tilde{z}) - \ell(g_{\Tilde{\mathcal{D}}^{\backslash i}}, \Tilde{z}) |\right]\leq 2 B\beta(g_{\Tilde{\mathcal{D}}})\gamma_p\frac{p}{m},
    \label{eq:weighted_bagging2}
\end{equation}

Here, the stability of bagging (subbagging) and the stability of the combiner are independent. The first multiplier on the right-hand side refers to the hypothesis stability of the combiner algorithm. It follows immediately that stacking improves the stability of bagging and subbagging by a factor proportional to the stability of the combiner algorithm. For instance, if we use $k$-NN as the combiner, which has hypothesis stability $k/m$, it improves the stability of bagging and subbagging by a factor proportional to $O(1/m)$, which, in theory, is a significant improvement.

To summarize, Equations~\eqref{eq:bag_stacking_2}~and~\eqref{eq:dag_stacking_2}, compared to Equation~\eqref{eq:stacking_hypothesis_stability}, imply that a larger number of base models $T$ improves the hypothesis stability of all three approaches: stacking, bag-stacking, and dag-stacking.
Conversely, bagging and subbagging improve the stability of the base learning algorithm, while adding a combiner on top improves it even further.

\section{Conclusion}
\label{sec:conclusion}
In this paper, we studied the effectiveness of stacking, bag-stacking, and dag-stacking from the perspective of algorithmic stability. This aspect allowed us to formally study the performance of stacking by analyzing its hypothesis stability and establishing a connection to bag-stacking and dag-stacking. Additionally, stacking turned out to improve stability by a factor of $O(1/m)$ when the combiner learning algorithm is stable, whereas subsampling/bootstrap sampling (within stacking) improved it even further. Finally, we found that the converse holds as well---stacking improves the stability of both subbagging and bagging.

\bibliographystyle{unsrt}

\end{document}